\documentclass{article}
\usepackage{PRIMEarxiv}
\usepackage{amsmath,amssymb,amsfonts}
\usepackage{algorithmic}
\usepackage{graphicx}
\usepackage{textcomp}
\usepackage{pgf-pie}
\usepackage{bm}
\usepackage{url}
\usepackage{caption}
\usepackage{pgf,tikz}
\usepackage{amsmath,amssymb,amsfonts}
\usepackage{algorithmic}
\usepackage{graphicx}
\usepackage{textcomp}
\usepackage{pgf-pie}
\usepackage{bm}
\usepackage{url}
\usepackage{caption}
\usepackage{pgf,tikz}
\usetikzlibrary{shapes.geometric, arrows, decorations.markings,positioning,calc}
\usepackage{enumerate,enumitem}

\usepackage{pgfplots}
\pgfplotsset{compat=1.17}
\usepackage{adjustbox}

\usepackage{tikz}
\usetikzlibrary{arrows.meta}

\def\innerradius{1.6cm}
\def\outerradius{2cm}

\newcommand{\wheelchartwithlegend}[1]{
  \pgfmathsetmacro{\totalnum}{0}
  \foreach \value/\colour/\name in {#1} {
      \pgfmathparse{\value+\totalnum}
      \global\let\totalnum=\pgfmathresult
  }

  \begin{tikzpicture}
    \pgfmathsetmacro{\wheelwidth}{\outerradius-\innerradius}
    \pgfmathsetmacro{\midradius}{(\outerradius+\innerradius)/2}
    \begin{scope}[rotate=90]
        \coordinate (L-0) at (\outerradius+5mm,-\outerradius-2.5cm);
        \pgfmathsetmacro{\cumnum}{0}
    \foreach [count=\i,remember=\i as \j (initially 0)] \value/\colour/\name in {#1} {
          \pgfmathsetmacro{\newcumnum}{\cumnum + \value/\totalnum*360}
            \pgfmathsetmacro{\percentage}{\value}
          \pgfmathsetmacro{\midangle}{-(\cumnum+\newcumnum)/2}
          \pgfmathparse{
             (-\midangle<180?"west":"east")
          } \edef\textanchor{\pgfmathresult}
          \pgfmathsetmacro\labelshiftdir{1-2*(-\midangle>180)}
          \fill[\colour] (-\cumnum:\outerradius) arc (-\cumnum:-(\newcumnum):\outerradius) --
          (-\newcumnum:\innerradius) arc (-\newcumnum:-(\cumnum):\innerradius) -- cycle;
          \draw  [Circle-,thin] node [append after command={(\midangle:\midradius pt) -- (\midangle:\outerradius + 1ex) -- (\tikzlastnode)}] at (\midangle:\outerradius + 1ex) [xshift=\labelshiftdir*0.5cm,inner sep=0pt, outer sep=0pt, ,anchor=\textanchor]{\pgfmathprintnumber{\percentage}\%};
          \node [anchor=north west,text width=5cm,font=\footnotesize] (L-\i) at (L-\j.south west) {\name};
          \fill [fill=\colour] ([xshift=-3pt,yshift=1mm]L-\i.north west) rectangle ++(-2mm,5mm);
          \global\let\cumnum=\newcumnum
      }
    \end{scope}
  \end{tikzpicture}
}
\tikzstyle{startstop} = [rectangle, rounded corners, 
minimum width=3cm, 
minimum height=1cm,
text centered, 
draw=black, 
fill=blue!30]

\tikzstyle{reject} = [trapezium, 
trapezium stretches=true, % A later addition
trapezium left angle=70, 
trapezium right angle=110, 
minimum width=3cm, 
minimum height=1cm, text centered, 
draw=black, fill=black!5!red!50]

\tikzstyle{select} = [rectangle, 
minimum width=3cm, 
minimum height=1cm, 
text centered, 
text width=3cm, 
draw=black, 
fill=green!30]

\tikzstyle{decision} = [diamond, 
minimum width=3cm, 
minimum height=1cm, 
text centered, 
draw=black, 
fill=violet!20]

\tikzstyle{arrow} = [thick,-latex]
\def\BibTeX{{\rm B\kern-.05em{\sc i\kern-.025em b}\kern-.08em
    T\kern-.1667em\lower.7ex\hbox{E}\kern-.125emX}}

\usepackage[utf8]{inputenc} % allow utf-8 input
\usepackage[T1]{fontenc}    % use 8-bit T1 fonts
\usepackage{hyperref}       % hyperlinks
\usepackage{url}            % simple URL typesetting
\usepackage{booktabs}       % professional-quality tables
\usepackage{amsfonts}       % blackboard math symbols
\usepackage{nicefrac}       % compact symbols for 1/2, etc.
\usepackage{microtype}      % microtypography
\usepackage{lipsum}
\usepackage{fancyhdr}       % header
\usepackage{graphicx}       % graphics
\graphicspath{{media/}}     % organize your images and other figures under media/ folder

\title{A Proposed Paradigm for Imputing Missing Multi-Sensor Data in the Healthcare Domain}

\author{
  Vaibhav Gupta \\
  Data Engineering\\
  Helmut-Schmidt-University \\
  Hamburg\\
  \texttt{guptav@hsu-hh.de} \\
  \And
  Florian Grensing\\
  Data Engineering\\
  Helmut-Schmidt-University \\
  Hamburg\\
  \texttt{grensinf@hsu-hh.de} \\
  \And
 Beyza Cinar \\
  Data Engineering\\
  Helmut-Schmidt-University \\
  Hamburg\\
  \texttt{cinarb@hsu-hh.de} \\
   \And
  Maria Maleshkova \\
  Data Engineering\\
  Helmut-Schmidt-University \\
  Hamburg\\
  \texttt{maleshkm@hsu-hh.de} \\
}
\begin{document}
\maketitle

\begin{abstract}
Chronic diseases such as diabetes pose significant management challenges, particularly due to the risk of complications like hypoglycemia, which require timely detection and intervention. Continuous health monitoring through wearable sensors offers a promising solution for early prediction of glycemic events. However, effective use of multisensor data is hindered by issues such as signal noise and frequent missing values. This study examines the limitations of existing datasets and emphasizes the temporal characteristics of key features relevant to hypoglycemia prediction. A comprehensive analysis of imputation techniques is conducted, focusing on those employed in state-of-the-art studies. Furthermore, imputation methods derived from machine learning and deep learning applications in other healthcare contexts are evaluated for their potential to address longer gaps in time-series data. Based on this analysis, a systematic paradigm is proposed, wherein imputation strategies are tailored to the nature of specific features and the duration of missing intervals. The review concludes by emphasizing the importance of investigating the temporal dynamics of individual features and the implementation of multiple, feature-specific imputation techniques to effectively address heterogeneous temporal patterns inherent in the data.
\end{abstract}


\section{Introduction} \label{sec:introduction}
Demographic changes and unhealthy lifestyles lead to a worldwide increase in patients with chronic diseases such as metabolic and coronary complications or neurological disorders. Chronic diseases can impede everyday activities and need lifelong medical care \cite{aditi1}. Thus, daily and, ideally, continuous monitoring is required to effectively prevent severe complications, forecast emergencies, and enable timely interventions. The emerging fields of sensors and wearables, as well as the Internet of Things (IoT) \cite{st}, have enabled the collection of physiological data in real-time \cite{rm}. In particular, the data integration of various sensors enhances the holistic representation of the health status and individual patient profiles.
Wearables and medical sensors can measure vital signs like the electrocardiogram (ECG), electroencephalogram (EEG), heart rate (HR), or accelerometers (ACC) non-invasively, whereas blood glucose (BG) levels require minimally invasive sensors. 
However, shortcomings of wearables include errors in assessment, device failures, battery exhaustion, and environmental factors, which can result in noise, unusable data samples, or missing data values.
Preprocessing, especially feature extraction and feature sampling to the same frequency, can also result in further missing values. Thus, the combination of heterogeneous sensors and data necessitates effective data engineering and imputation methods.
Treating missing data values is crucial for improving the overall data quality and the subsequent performance of the applied machine learning models.

One increasingly prominent and fast-progressing chronic disease is diabetes \cite{American Diabetes Association}. Patients with diabetes have a disrupted glucose metabolism, leading to glucose levels greater than 180 mg/dL (hyperglycemic range) or less than 70 mg/dl (hypoglycemic range) due to improper insulin dosage \cite{American Diabetes Association, mayo}. If glucose levels are not monitored frequently, conditions like hypoglycemia, which is dealt mainly with T1D patients, can lead to dangerous complications such as dizziness, nausea, or lightheartedness.

This review examines the preprocessing of datasets to enhance the early detection of hypoglycemia in patients with type 1 diabetes. Data from various sensors can be integrated into a single dataset to aid in hypoglycemia prevention. As previously mentioned, these unified datasets often face numerous data quality issues. Consequently, they must undergo proper preprocessing techniques before being utilised as input for ML models, as highlighted in studies \cite{b1,b13,b8,b6}. 
This review evaluates studies predicting hypoglycemia and compiles the relevant datasets involved. We examine the limitations of these datasets with regard to data quality. In addition, the features of the datasets, including their temporal behavior, are comprehensively analysed. Furthermore, we study prominent methods used to address missing values. The combined data often includes missing values across variables (cross-sectional data) and over time (longitudinal data), necessitating thorough preprocessing. Addressing missing values in hypoglycemia prediction is crucial, as most available datasets contain only a limited number of data samples. Consequently, missing data and poor data quality significantly reduce both the overall quality and quantity of data when combining information from various sensors.

This systematic review investigates different approaches and imputation methods for treating missing values in time series data while addressing the following research questions:
\begin{enumerate}
\item  What datasets and sensors are used for the prediction of hypoglycemia?
\item What are the different features used for the prediction of hypoglycemia and what is their behavior with time?
\item What different imputation techniques are utilised for hypoglycemia prediction?
\item What imputation techniques from other healthcare domains can be adopted for the prediction of hypoglycemia?
\end{enumerate}

Based on the reviewed studies, we conduct a quantitative analysis of various imputation techniques, discussing their strengths and weaknesses. In our analysis, we identify research gaps related to these imputation techniques and explore machine learning and deep learning methods from other healthcare domains to address longer missing values. Consequently, we suggest a paradigm based on the reviewed studies that involves imputing different features separately, taking into account varying time gaps and analysing a range of imputation techniques. However, a universal imputation technique optimized for specific time gap lengths has yet to be established. Moreover, the accuracy and reliability of hypoglycemia prediction are influenced not only by the imputation strategy, but also by the underlying characteristics of the dataset and the machine learning models employed \cite{b30}. The proposed paradigm, though developed in the context of studies focused on hypoglycemia prediction, is designed to be adaptable to a wide range of healthcare applications that rely on sensor-based data and face inherent data quality limitations. 

The main contributions of this paper are threefold: Firstly, it analyses the behavior of different features involved in hypoglycemia prediction over time. Secondly, it proposes a paradigm for imputing missing values for feature gaps of various lengths. Lastly, it lists machine-learning and deep learning techniques from other healthcare domains that could be adopted to predict hypoglycemia with larger time gaps. 

The outline of the paper is as follows: Section \ref{sec:methodology} describes the methodology used to select the studies for the review paper. Section \ref{sec:datasets} describes the datasets and sensors utilised in studies for the prediction of hypoglycemia. Section \ref{sec:data analysis} investigates important features and their behavior with time in the context of hypoglycemic studies. Following this, \ref{sec:imputation} comprises three subsections. First, subsection \ref{sec:imputation3} comprises of data imputation techniques used in hypoglycemic prediction studies, the following subsection \ref{sec:analysis}  analyses the imputation techniques used, and the last subsection \ref{sec:imputation4} describes the machine learning and deep learning imputation techniques used in time series healthcare datasets. Finally, section \ref{sec:results} answers the research questions followed by the conclusion.

\section{Materials and Methods} \label{sec:methodology}

This review aims to identify effective preprocessing and imputation methods for sensor data within the healthcare domain. It focuses on studies related to features used in hypoglycemia prediction or publications emphasizing diabetes. The review is divided into three parts. First, datasets and features used in diabetes and hypoglycemia research are identified. With this foundation, necessary preprocessing steps and imputation techniques are investigated and analysed, extracted from studies predicting hypoglycemia, comprising the second part. Similarly, the third part consists of machine learning and deep learning imputation techniques investigated from other healthcare domains working with time series data. Finally, we propose our results for handling missing values for specific gaps of separate features in hypoglycemia prediction.
The following section describes the methodology used to search and select various studies included in different parts of this literature review.

\subsection{Search Process}
The systematic literature review was based on a step-by-step process starting from the aim of the study, inclusion, and exclusion criteria as described in table \ref{table1}.
The literature review was conducted during four weeks, from May 1 to 28, 2024. 

\begin{table}[h]
    \caption{Study Selection Criterion}
    \label{table1}
    \setlength{\tabcolsep}{3pt}
    \begin{tabular}{|p{55pt}|p{355pt}|}
        \hline
        \textbf{Aim} & This review aims to study various imputation techniques for filling missing time series data in the healthcare domain. \\
        \hline
        \textbf{Inclusion Criteria} & 
        \begin{itemize}
            \item The publications using sensors and datasets for predicting hypoglycemia in T1DM patients.
            \item The publications on hypoglycemia that address the problem of preprocessing and imputing missing values in sensor data.
            \item The publications that include imputation techniques (machine learning and deep learning) in time series healthcare data.
            \item The publications that include analysis of different features for the prediction of hypoglycemia.
            \item The publications that use the prediction models for the prediction of hypoglycemia.
            \item The publications that are published after 2018.
    \end{itemize}  \\
    \hline
    \textbf{Exclusion Criteria} & 
    \begin{itemize}
     \item The publications on hypoglycemia not including imputation methods for missing values.
    \item The publications that give only limited or unclear explanation of the used imputation technique.
    \item Duplication of imputation techniques used in studies like Linear Interpolation etc.
    \end{itemize}\\
    \hline
    \end{tabular}
\end{table}

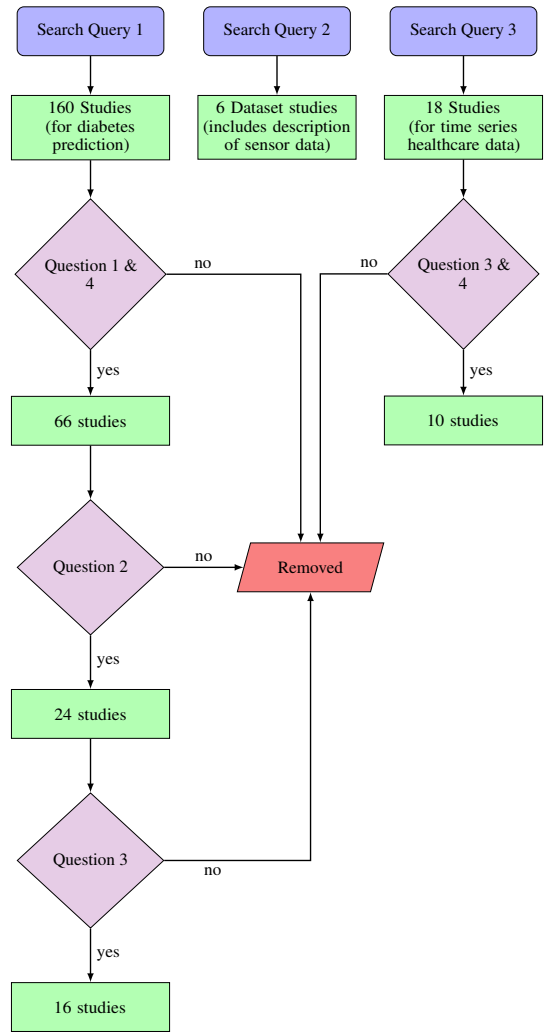
\begin{figure}
\begin{center}
\vspace{0.5cm}
\scalebox{0.65}{%
\begin{tikzpicture}[node distance=1.5cm]

\node (start) [startstop] {Search Query 2};
\node (start1) [startstop, left = 0.8cm of start] {Search Query 1};
\node (start2) [startstop, right = 0.8cm of start] {Search Query 3};

\node (select8) [select, below =0.8cm of start] {6 Dataset studies \\ (includes description of sensor data)};
\node (select48) [select, below =0.8cm of start1] {160 Studies \\ (for diabetes prediction)};
\node (ic1) [decision, below of=select48, yshift=-1.5cm] {\parbox{2cm}{\centering Question 1 \& 4 }};
\node (select32) [select, below of=ic1, yshift=-1.5cm] {66 studies};
\node (ic2) [decision, below of=select32, yshift=-1.5cm] {\parbox{2cm}{\centering Question 2}};
\node (select22) [select, below of=ic2, yshift=-1.5cm] {24 studies};
\node (ic_1) [decision, below of=select22, yshift=-1.5cm] {\parbox{2cm}{\centering Question 3}};
\node (select13) [select, below of=ic_1, yshift=-1.5cm] {16 studies};

\node (select18) [select, below =0.8cm of start2] {18 Studies \\ (for time series healthcare data)};
\node (all) [decision, below of=select18, yshift=-1.5cm] {\parbox{2cm}{\centering Question 3 \& 4}};
\node (select10) [select, below of=all, yshift=-1.5cm] {10 studies};

\node (reject1) [reject, right of=ic2, xshift=3cm] {Removed};

\draw [arrow] (start) -- (select8);
\draw [arrow] (start1) -- (select48);
\draw [arrow] (start2) -- (select18);
\draw [arrow] (select48) -- (ic1);
\draw [arrow] (ic1) -- node[anchor=west] {yes} (select32);
\draw [arrow] (select32) -- (ic2);
\draw [arrow] (ic2) -- node[anchor=west] {yes} (select22);
\draw [arrow] (select22) -- (ic_1);
\draw [arrow] (ic_1) -- node[anchor=west] {yes} (select13);

\draw [arrow] (select18) -- (all);
\draw [arrow] (all) -- node[anchor=west] {yes} (select10);

\draw [arrow] (all) -| node[anchor=south, xshift=1cm] {no} ($(reject1.north)+(0.2cm,0)$);
\draw [arrow] (ic1) -| node[anchor=south, xshift=-2cm] {no} ($(reject1.north)-(0.2cm,0)$);
\draw [arrow] (ic2) -- node[anchor=south] {no} (reject1);
\draw [arrow] (ic_1) -| node[anchor=north, xshift=-2cm] {no} (reject1);
\end{tikzpicture}}
\end{center}
\caption{Flowchart  of selection process}
\label{fig:flow}
    
\end{figure}

The clinical datasets used for prediction of hypoglycemia were searched on "Google Scholar" with key terms like "clinical datasets", "diabetes prediction datasets" which were then shortlisted if the description of sensor data was included in the dataset. A total of 6 publications were selected from these keywords (Search Query 2 in figure \ref{fig:flow}).

Literature, which includes preprocessing and imputation techniques related to hypoglycemia prediction, was searched in the "IEEE Xplore" and "Scopus" databases. The first search query was: ("sensors" OR "datasets") AND ("hypoglycemia" OR "diabetes") AND ("Type 1") AND ("missing" OR "imputation" OR "preprocessing") AND ("machine learning" OR "deep learning"), which returned 6 results from "Scopus." Then, to increase the number of results, the search query was modified to: ("sensors" OR "datasets") AND ("hypoglycemia" OR "diabetes") AND ("Type 1") AND ("machine learning" OR "deep learning"), which outputted 114 papers on "IEEE Xplore" and 20 on "Scopus." In the end, different combinations of strings like "hypoglycemia," "imputation," and "preprocessing" were searched in Google Scholar, and 20 publications were selected.
Both the search queries and keywords combined (Search Query 1 in figure \ref{fig:flow}) resulted in a total selection of 160 studies. The second search query was used as it produced more target-specific publications than the first one. From the first search query, the conclusion is drawn that most publications used to predict hypoglycemia do not focus on preprocessing and imputation techniques.

The third search query for searching the publications in "IEEE Xplore" and "Scopus," related to imputation techniques used in the time series healthcare domain was: ("clinical data") AND ("time series data") AND ("imputation methods") and 3 and 1 publications were found, respectively. In the end, key terms like "imputation," "time series datasets," and "healthcare domain" were searched in "Google Scholar," and 14 publications were selected. Thus, 18 publications were selected from this search query (Search Query 3 in figure \ref{fig:flow}).

The flowchart of the selection process illustrates how the studies are split into three sections. To narrow down the selection criteria, the following questions were used:

\begin{enumerate}
    \item Question 1: Is a description of sensor data included?
    \item Question 2: Is a description of imputation techniques for datasets included?
    \item Question 3: Is a comprehensive explanation of imputation techniques given?
    \item Question 4: Is a new imputation technique used, which has not been applied in other studies?
\end{enumerate}

Publications from all sections went through a thorough analysis during the literature review and the results, according to our research questions, are discussed in the following section.

\section{Results}

This section presents the results of the literature review. First, the datasets and feature dynamics utilised in diabetes research and hypoglycemia prediction over time are examined. Second, preprocessing steps and imputation techniques used in diabetes research are discussed. Moreover, an analysis of imputation strategies is conducted. Finally, the applied machine learning and deep learning methods that can fill longer time gaps are summarized, followed by a discussion.

\subsection{Datasets Used to Predict Hypoglycemia} \label{sec:datasets}

This section describes the datasets and sensors used to predict hypoglycemia found during the search, as described in section \ref{sec:methodology}. 

According to their acquisition procedures, the datasets can be categorized as clinical and simulator-based.
Clinical datasets collect data from subjects in hospitals or living in normal conditions. In contrast, simulator-based datasets are generated by simulators, which are approved by the Food and Drug Administration (FDA) for insulin trials.  

\begin{figure}[h!]
    \centering
    \begin{tikzpicture}
        \pie[
            color={red, blue!50,magenta, green, yellow},
            text=legend,
            radius=2
        ]{
            65/OhioT1DM, 
            10/ABCD4, 
            10/Replace BG,
            5/D1NAMO,
           10 /DI Advisor
        }
    \end{tikzpicture}
    \caption{Clinical Datasets used in studies}
    \label{fig1:piechart}
\end{figure}
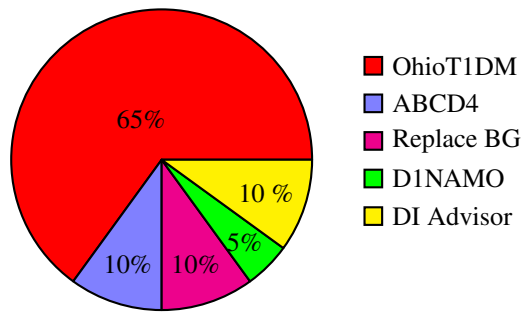

Table \ref{table2} presents available clinical datasets utilised for hypoglycemia prediction. Five out of the six datasets listed in the table are utilised in the reviewed publications, with the OhioT1DM being the most commonly used one (65\%) (See figure \ref{fig1:piechart}). The detailed description of the datasets are being discussed in the following subsection.

\subsubsection{Dataset Description}

\begin{enumerate}
    \item \textbf{OhioT1DM Data set:} Sensors used to accumulate the readings are Medtronic Enlite CGM, Basis Peak fitness bands and  Empatica Embrace. The dataset includes 12 patients and 8 weeks of readings The variables measured by sensors includes following information: CGM BG level after every 5 minutes; BG levels from periodic self-monitoring of blood glucose (finger sticks); insulin doses, both bolus and basal; self-reported meal times with carbohydrate estimates; self-reported times of exercise, sleep,  work, stress; illness  data from the Basis Peak or Empatica Embrace band.
     
      \item \textbf{D1NAMO Data set:} Sensors used for accumulating the readings are Zephyr Bioharness 3 and iPro2 Professional CGM sensor. The dataset includes data of 20 healthy patients and 9 T1D patient. The features included are  Breathing Rate; Accelerometer Signals; Glucose measurement every 5 minutes; ECG data.
   
    \item  \textbf{Replace BG Data set:} The CGM device used in the study is the Dexcom G4 Platinum Continuous Glucose Monitoring System with modified algorithm. For CGM+BGM  Dexcom G4 Platinum Continuous Glucose Monitoring System with modified algorithm + Abbot Precision Xtra Blood Glucose-Ketone Meter is used. The trial period is of 26 weeks. Participants are randomly assigned 2:1 to the CGM-only (n = 149) or CGM+BGM (n = 77) group. The outcome features available are: Percentage of Time in Range of 70 to 180 mg/dl, measured With CGM; Mean Glucose; Measurement of Glycemic Variabilty: Coefficent of Variation; Change in HbA1c; Percentage of Time With Sensor Values, number of participants with no worsening of HbA1c with different ranges; number of participants with severe hypoglycemia events in different ranges.
  
    \item \textbf{DI Advisor Data set:} Sensors used in this study is Dexcom G5 Mobile CGM, Accu-Check and Abbot Freestyle. In the research project DI Advisor, data is collected from 59 T1D patients (37 males/22 females) participating in a three-day in-hospital study. Seven features are considered from this data set: interstitial glucose (mg/dL) measured by Abbott Freestyle; self reported insulin intakes (U) for basal, bolus, and correction doses; and meal nutrients content (mg) for CHO, protein, and lipids.
   
    \item \textbf{ABCD4 Project Data set:}  Sensor used in study is Dexcom CGM devices.  10 T1D diabetic subjects participated and were monitored for 6 consecutive months. Information on meals and insulin dosages was recorded by the patients themselves through a dedicated app.
    
    \item \textbf{HUPA-UCM Diabetes Data set:} CGM data was acquired by FreeStyle Libre 2 CGMs, and Fitbit Ionic smartwatches were used to obtain steps, calories, heart rate, and sleep data for at least 14 days.  The data is acquired from 25 people with T1D. This dataset provides a collection of CGM data, insulin dose administration, meal ingestion counted in carbohydrate grams, steps, calories burned, heart rate, and sleep quality and quantity assessment. Dataset is not considered in the studies included in our research on the basis of our exclusion criteria.
\end{enumerate}
\begin{table}[h]
\caption{Clinical Datasets used in different studies}
\label{table2}
\setlength{\tabcolsep}{3pt}
\begin{tabular}{|p{85pt}|p{225pt}|p{85pt}|}
\hline 
\textbf{Dataset} & \textbf{Sensors Used in Study} & \textbf{Considered in study}  \\
\hline
 OhioT1DM \cite{OHIO}& Medtronic Enlite CGM, Basis Peak fitness band, Empatica Embrace & YES \\
\hline
D1NAMO \cite{D1NAMO}& iPro2 Professional CGM, Zephyr Bioharness 3 &
 YES \\
\hline
Replace-BG Data set  \cite{Replace}& Dexcom G4 Platinum, Abbot Precision Xtra Blood Glucose-Ketone
Meter, Accu Check & YES \\
\hline 
DI Advisor dataset \cite{DIadvisor}& Dexcom G5 Mobile CGM, Accu Check, Abbott Freestyle & YES  \\
\hline 
ABC4D project \cite{ABCD} & Dexcom CGM & YES \\
\hline
HUPA-UCM Diabetes Dataset \cite{hupa} & FreeStyle Libre
2  & NO \\
\hline
\end{tabular}
\end{table}

The analysis of the datasets shows that the features' frequency levels are different throughout the datasets. The CGM BG values are usually collected over 5 minutes, and other features are collected according to the physiological sensors used to collect them. 
For example, in the OhioT1DM dataset \cite{OHIO}, the data for individuals who wore the Basis Peak band includes 5-minute aggregations of heart rate, galvanic skin response (GSR), skin temperature, air temperature, and step count, In contrast, the data for those who wore the Empatica Embrace band includes 1-minute aggregations of GSR, skin temperature, and magnitude of acceleration.\\

Through the analysis of the clinical datasets, some challenges related to the data could be identified.
First, the limited quantity of data is evident from the above description of datasets.
The datasets are relatively small when considering the tasks of generalisation of features and training the machine learning models, which can result in over-fitting. Second, through the feature analysis of the datasets, missing values of the key features at different time intervals could be identified, which can occur due to human or technical error. This could be observed in almost all the datasets in tables \ref{table2}. The incomplete values may also result from features having varying frequencies and from attempts to sample features to the same frequency level through the use of interpolation or extrapolation techniques. Third, outliers could be seen in datasets due to synchronization errors. For instance, outlier values in the D1NAMO dataset \cite{D1NAMO} occur in the HRV feature.
Lastly, the diabetes datasets comprise a disproportionate number of hypoglycemia cases, as seen in all the datasets listed in table \ref{table2}. As a result, biased machine learning models could perform poorly when predicting minority classes.

To overcome the challenges, some studies \cite{b11},\cite{b2} apply virtual datasets, generated by simulators. 
Simulator-based datasets are virtual datasets, which in recent times are used by research studies \cite{b11}, \cite{b2} to train machine learning models due to the limitation of small data volumes of existing clinical datasets. 

Zhu \textit{et al.} \cite{b11}, and Li \textit{et al.} \cite{b2} generate ten virtual subjects with T1D with the UVA/Padova T1D simulator. The FDA approves this simulator for insulin trials.

Although the simulator dataset offers higher data quality due to controlled conditions and fewer missing values, it cannot be directly compared or substituted for clinical datasets. This is primarily because simulator-based datasets do not capture the natural variability in real-world clinical environments, such as sensor noise resulting from patient movement, respiration, or other physiological activities. Consequently, models trained solely on simulator data may not generalize well to clinical scenarios. To address this limitation, simulator-based datasets are best utilised within a transfer learning framework—where a model is initially trained on the simulator data to learn foundational patterns and then fine-tuned using a relatively smaller amount of clinical data.  

Consequently, we do not further consider simulator-based datasets in this review, as they do not reflect the challenges of real sensor data, and do not necessarily involve preprocessing and imputation techniques. To critically assess the quality of the datasets, individual features must be examined, as elaborated in the subsequent subsection. 

\subsection{Features Related to Glucose Levels}
\label{sec:data analysis}

Wearables are used to collect various features at different time intervals from the presented datasets. This section focuses on elucidating the factors influencing the variability of features used to predict hypoglycemia, as the behavior of these features depends on multiple factors.
Effective recognition of feature patterns is essential, as preprocessing procedures and imputation techniques depend on their dimensions and complexity.

In the studies \cite{b12}, \cite{b9}, \cite{b4} used for predicting hypoglycemia, the commonly used features are the blood glucose values supplemented with insulin and meal data. Datasets in recent studies are augmented with other physiological signals than just glucose to improve the accuracy of their prediction models. Datasets such as the ABCD4 project \cite{ABCD}, the DI Advisor \cite{DIadvisor}, and the Replace BG \cite{Replace} do not include data of physiological signals like ECG, EEG and accelerometer signals but the D1NAMO \cite{D1NAMO}, OhioT1DM \cite{OHIO} and HUPA-UCM Diabetes datasets \cite{hupa} use glucose values supplemented by ECG, EEG and accelerometer signals. Studies are using ECG, EEG, and accelerometer signals since, biologically and mathematically, a correlation between BG values and other physiological signals has been shown. The change in ECG values can depend on the change in blood glucose values, out of many factors impacting it, as mentioned in the Li \textit{et al} \cite{b2} paper. Similarly, the change in accelerometer signals could impact BG values \cite{acc}. ECG signals are processed, and advanced features like Heart Rate (HR) and Breathing Rate (BR) are extracted from them. These features are used in predictive ML models. The most common feature among the ECG features used in studies is HR values. Thus, the prominent features used in studies to predict hypoglycemia are glucose readings, HR values, and accelerometer readings.  

Using the features in the studies requires a thorough understanding of their temporal behavior. A specific feature may experience a sharp increase or abrupt decrease, and it is important to comprehend the underlying cause and factors causing this.

Among the factors considered for predicting hypoglycemia, the initial feature of interest is blood glucose levels.
Some variables, including insulin consumption, intense exercise, and meal intake, can cause a quick change in blood glucose levels. This causes BG readings to alter suddenly, as seen in the clinical datasets used for this investigation. When fasting, blood glucose levels fluctuate more slowly than when insulin is taken, when exercising vigorously, and when a meal is consumed. The BG values can rise from 90 mg/dL to 140 mg/dL after fasting in the span of half an hour and can reduce from 170 mg/dL to 90 mg/dL in 15 minutes, as observed in the datasets \cite{D1NAMO, OHIO}.

The next feature of interest is the HR values that depend upon several factors like stress, illness, etc. The heart value fluctuates rapidly over time; for example, it changes from 60 bpm to 140 bpm in 10 minutes and can drop again from 180 bpm to 100 bpm in the next couple of minutes as observed in datasets \cite{D1NAMO}, \cite{OHIO}. 

The range of values in accelerometer signals varies from -1 to 1. The rate of change of the accelerometer depends upon several factors, such as arm motion, physical exercise, etc. These values can also suddenly change if there is a sudden movement of the body.

EEG is another signal which shows a correlation between hypoglycemia and brain waves. Regarding the correlation between hypoglycemia and EEG, since the EEG is directly related to the metabolism of brain cells, a failure of cerebral glucose supply can cause early changes in EEG signals. These changes could affect the metabolism of glucose values. As no concrete methods are available for dealing with missing values in EEG data, these papers \cite{snogdal2012detection} are not included in our table.

While biological studies have provided insights into the reasons behind the correlation among these features and their interdependencies, statistical analyses have yet to fully investigate the dynamics of how these features change over time.
To maximize the analytical value of the dataset, thorough feature preprocessing is essential. The specific preprocessing steps are outlined in detail in the following subsection.

\subsection{Preprocessing and Imputation} \label{sec:imputation}

Data preprocessing transforms raw data into a format that can be effectively used for machine learning algorithms. The preprocessing techniques are essential in the context of our review but some preprocessing techniques could lead to missing values in the data. In particular, extracting or sampling the features on the same frequency level could result in missing values. 

To impute the missing values, naturally inherent in the data and created by preprocessing techniques are reviewed in the following subsection. Further a comprehensive analysis of the imputation techniques employed in the studies  are presented. 

\subsubsection{Data Imputation Techniques used in Studies} \label{sec:imputation3}

This section describes the methods applied to missing values in studies predicting hypoglycemia. The missing values can be categorized differently based on their occurrence, proportion and time gaps of different lengths. Alam \textit{et al.} \cite{alam2023} categorize them based on occurrence into three types: Missing Completely at Random (MCAR), Missing at Random (MAR), and Missing Not at Random (MNAR). When the missing values occur without any pattern, they are called MCAR. Missing values occurring related to other observed values are termed MAR. In the case of MNAR, the missing values are directly related to the values themselves.

Missing values can be classified based on their proportion within a dataset. When the proportion of missing values is less than 1\%, they are considered trivial \cite{alam2023}. The proportions can vary depending on the volume of the dataset and can be categorized into several ranges: 1\% to 5\%, 15\% to 30\%, 30\% to 50\%, and more than 50\%. Distinguishing between these categories of missing values is essential for determining effective strategies to address them. For example, different proportions may require varied approaches to handle missing data, which will be discussed in this section.

Another approach to classifying missing values is based on the duration of the gaps, such as gaps of 5 minutes, 10 minutes, 15 minutes, less than 30 minutes, or up to 2 hours, as applied in the analysis of the DiaData dataset \cite{cinar2022}. This classification is important because gaps of different durations may necessitate distinct strategies to ensure appropriate handling and maintain data integrity.

One way to deal with missing values is to delete the rows containing a missing value for a particular feature \cite{b5,b8}. However, the removal of missing data may lead to a reduction in statistical power. 
This method is not the most suitable to implement, as it addresses an existing limitation in the quality of the data in clinical studies, as outlined in the challenges presented in section \ref{sec:datasets}. Additionally, some studies, such as the one by Cinar \textit{et al.} \cite{beyza}, handle missing values by replacing them with zeros to maintain uniformity in the dataset.

A proposed approach to addressing missing values involves using imputation techniques. These techniques can reduce statistical loss but may also introduce bias into the datasets. Tables \ref{table4}, \ref{table5}, and \ref{table6} outline the various data imputation methods used in different studies. 

The studies are categorized into three criteria: 
\begin{enumerate}

\item Table \ref{table4} lists studies that utilise single imputation methods for blood glucose (BG) values, applied to different time intervals while excluding certain small or large gaps based on specific preferences.
  
\item Table \ref{table5} specifies studies that impute BG or physiological values.
  
\item Table \ref{table6} encompasses two distinct categories of studies: those that use the same imputation methods to fill missing gaps in blood glucose (BG) data but apply different techniques for gaps of varying lengths, and those that employ a combination of imputation techniques to fill in BG values or other features outlined in the dataset.
\end{enumerate}

The studies in table \ref{table4} employ K Nearest Neighbour (KNN)-based imputation, Cubic Interpolation, Mean, Linear Imputation, and Spline Techniques. Studies in table \ref{table5} adopt Mean, KNN, Forward Fill, and Linear imputation techniques to fill BG or other physiological values (heart rate, calories burnt, skin temperature, etc). However, using the same imputation technique for both BG and physiological signals, as done by Mantena \textit{et al.} \cite{b10}, is not recommended as there is uncertainty in the correlations of both features concerning time. Some studies, as listed in table \ref{table6}, leverage multiple imputation techniques for varying gaps for the blood glucose feature or use a combination of imputation techniques for different features to predict BG values.

\begin{table}[h]
\caption{Imputation techniques used for BG values}
\label{table4}
\setlength{\tabcolsep}{3pt}
\begin{tabular}{|p{55pt}|p{45pt}|p{245pt}|}
\hline 
\textbf{Dataset} & \textbf{Year}  &\textbf{Imputation Techniques} \\
\hline

 Zhu \textit{et al.} \cite{b11} & 2020 &  Linear Interpolation + median filter (Training dataset) + Extrapolation (Testing dataset) 
\\
\hline
  K.Li \textit{et al.} \cite{b6}& 2020 & Spline interpolation or extrapolation techniques are used when the missing CGM data are less than one hour (12 samples). \\
\hline
  M. Jaloli and M. Cescon \cite{b4}& 2022 &  Missing BG Datapoints less than 60 minutes were linearly interpolated. No interpolation for values greater than 60 minutes.
\\

\hline
Berikov \textit{et al.} \cite{b12} & 2022 & BG gaps for more than 30 minutes were excluded. Shorter intervals of missing values were linearly extrapolated based on surrounding observations.
\\
\hline
Faccioli \textit{et al.} \cite{b9} & 2023 &  Large gap: discarded. For training data: cubic splines. For test data: a first-order polynomial is fitted. \\

\hline 
\end{tabular}
\end{table}

\begin{table}[h]
\caption{Imputation Technique Used for BG or Physiological Values}
\label{table5}
\setlength{\tabcolsep}{3pt}
\begin{tabular}{|p{55pt}|p{45pt}|p{245pt}|}
\hline 
\textbf{Dataset} & \textbf{Year}  &\textbf{Techniques Used} \\
\hline
Vahedi \textit{et al.} \cite{b7} & 2018 & BG value is imputed by taking the closest preceding value. Physiological features are imputed with personalized mean values.
 \\

\hline
Mantena \textit{et al.} \cite{b10} & 2021 & Predictors which had missing entries over 40\% were removed; Missing data for all other features were determined using KNN-based imputation. \\
\hline
 
 Li \textit{et al.} \cite{b2} & 2021 & ECG outliers are replaced by adjacent value average and the BG values linearly interpolated. 
\\
\hline
Dave \textit{et al.} \cite{b1} & 2022 & ECG values are filtered and the discarded segment is filled by average, no data cleaning for accelerometer signals.  
\\
\hline
 \end{tabular}
\end{table}

\begin{table}[h]
\caption{Multiple imputation Techniques used}
\label{table6}
\setlength{\tabcolsep}{3pt}
\begin{tabular}{|p{55pt}|p{45pt}|p{245pt}|}
\hline 
\textbf{Dataset} & \textbf{Year}  &\textbf{Multiple Techniques Used} \\
\hline
Jeon \textit{et al.} \cite{b26}& 2019 &  Forward Fill, Spline Interpolation, Stineman Interpolation, Kalman Smoothing, Linear Interpolation \cite{b26} and Emperical imputation method (Emp).
 \\
 \hline
 Md Shafiqul, Qaraqe and Belhaouari \cite{b27} & 2021 & Linear Interpolation for one missing BG value, Nearest Neighbor for more than two missing BG values, same data of previous day for missing values of 1 day and ARMA for missing BG values of 2 or more than 2 days \\

\hline
Butt, Khosa and Iftikhar \cite{b29}& 2023 & Forward fill for one missing BG value and Linear Interpolation for more than two missing BG value
 \\
\hline
Acuna, Aparicio and Palomino \cite{b28} & 2023 &  Hourly mean, linear interpolation, spline interpolation and polynomial interpolation, Kalman Smoothing for BG and meal intake values
\\\hline

\end{tabular}
\end{table}

Butt, Khosa, and Iftikhar \cite{b29} work on the OhioT1DM datasets. The forward fill rule is employed to fill in the next missing BG value. Several interpolation techniques are compared for two or more consecutive missing BG values, including linear, Akima, cubic, spline, and shape-preserving methods. They report that linear interpolation is the best technique. They use a multi-layered long short-term memory (LSTM)-based recurrent neural network to predict blood glucose levels in patients with type 1 diabetes.

Acuna, Aparicio, and Palomino \cite{b28} deploy the OhioT1DM dataset and use CGM and meal intake features for their results. They use four imputation techniques, including hourly mean, linear interpolation, spline interpolation, and polynomial interpolation, as well as two smoothing techniques: Kalman Smoothing and Polynomial Smoothing. To test and compare the trends of imputation techniques, they utilise a specific time frame for one subject and give their conclusions based on all subjects of the data. They use XGboost, 1DCNN, and transformer models with Root Mean Square Error (RMSE) values as the metric to test the performance of models using the listed imputation techniques.

They combine imputation techniques with the models and report that the impact of the preprocessing methods depends on the machine learning model utilised. 
When only CGM is used and no missing values are interpolated, the transformer model performs the best. When only CGM is used as a predictor, linear interpolation is the best imputation method. Using only CGM as a predictor, Kalman smoothing yields better results than smoothing splines for the Transformer and XGBoost algorithms, but smoothing splines perform better for 1D-CNN. Applying imputation, including a second feature (meal), increased the RMSE of the deep learning prediction models, and only the XGBoost algorithm was unaffected.

Md Shafiqul, Qaraqe, and Belhaouari \cite{b27} apply multiple imputation techniques concerning the missing data of BG values. If there is one missing value, the previous and next values are filled by Linear Interpolation. The nearest neighbors are applied if there are more than two consecutive missing values. The missing values are interpolated by taking the average of the eight nearest neighbors. For missing values of more than 24 hours, the previous day's trend fills the value; a single Autoregressive Moving Average (ARMA) is utilised for over two days.

Jeon \textit{et al.} \cite{b26} compare 16 imputation techniques \cite{b26} for their feature set. The chosen prediction model for forecasting BG values is XGBoost. RMSE, Mean Absolute Error (MAE), and Pearson Correlation Coefficient (PCC) are computed for the performance evaluation of predictive models. Out of the 16 imputation methods, five imputation methods, namely Forward Fill, Spline Interpolation, Stineman Interpolation, Kalman Smoothing with structural models, and Linear Interpolation, are selected by using a cluster map between imputation methods and performance metrics, and last, using the rank product. In addition, they formulate the empirical imputation method (Emp). In this method, they collect values from the training set observed within ±5 minutes of the missing value’s timestamp for each missing value timestamp. Given this temporally constrained empirical distribution of feature values, they impute the missing value with either the mean of the distribution (Emp-mean) or a randomly chosen value (Emp-random) from the distribution.
They conclude that their ensemble models, which aggregated predictions from all five models, showed an approximately 10 percent improvement in RMSE in realistic settings over any model alone. 

Rehman \textit{et al.} \cite{b30} analysis of prediction of postprandial hypoglycemia relies on simulator data sets, which do not align with our study. Therefore, the results are not extensively discussed; only the new interpolation techniques are emphasized here.

Piecewise Cubic Hermite Interpolating Polynomial (PCHIP) interpolation technique is particularly useful for preserving trends in the data and avoiding overshooting. The mathematical equation for this is:
\begin{equation}
y(x)=\sum_{i=1}^{n-1}(a_i(x-x_i)^3+b_i(x-x_i)^2+c_i(x-x_i)+d_i)
\end{equation}
where $(x_i, y_i)$ are the known data samples, and $n$ is the number of data samples. The coefficients $a_i, b_i, c_i$, and $d_i$ are determined such that the slope at each data sample matches the slope of the cubic spline interpolating the data. The other technique is Makima interpolation which effectively reduces oscillations and improves accuracy in highly fluctuating datasets. The mathematical equation for this is:
\begin{equation}
y(x)=\sum_{i=1}^{n-1}(w_i^3(x-x_i)^3+w_i^2(x-x_i)^2+w_i(x-x_i)+1) y_i
\end{equation}

where $(x_i, y_i)$ are the known data samples, and n is the number of data samples. The weights $w_i$ are calculated based on the slopes of the data samples, with a weighted average used to determine the interpolation polynomial at each data sample.

The reviewed studies used several interpolation techniques. In the following subsection, we evaluate the efficacy of the imputation strategies employed and provide a quantitative summary of the imputation techniques.

\subsubsection {Analysis of Imputation Techniques} 
\label{sec:analysis}
This section analysis imputation techniques used in studies predicting hypoglycemia and examines their applications in addressing missing values. In the initial part, we quantify the imputation strategies used in all features according to their frequency of implementation accompanied by their descriptions. Next, we categorize the imputation techniques based on varying time gaps. Lastly, we discuss the potential explanations for using particular imputation techniques for a certain time gap. 

\begin{figure}[h!]
\begin{tikzpicture}
    \begin{axis}[
        ybar,
        symbolic x coords={LI, SI, KNN, Mean, FF, Stineman, KS, PI, ARMA},
        xtick=data,
        xlabel={Imputation Techniques},
        ylabel={Number of Studies},
        nodes near coords,
        enlarge x limits=0.15,
        ymin=0,
        bar width=0.5cm,
        x tick label style={rotate=45, anchor=east},
    ]
    \addplot coordinates {
        (LI,9) (SI,4) (KNN,2) (Mean,2) (FF,2) 
        (Stineman,1) (KS,1) (PI,1) (ARMA,1)
    };
    \end{axis}
\end{tikzpicture}
\caption{Types of Imputation Techniques Used}
    \label{fig4:barchart}
    \end{figure}
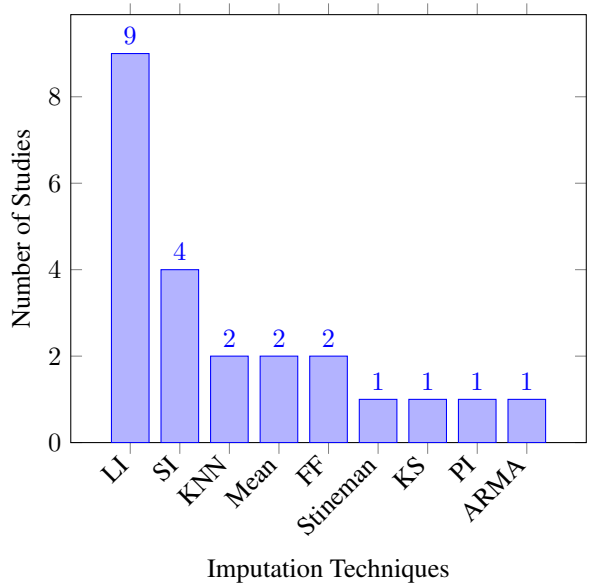

Imputation techniques can be classified into two parts: Statistical and Machine Learning. Statistical techniques use descriptive statistical tools and equations, whereas ML techniques use predictive models to impute the missing values.
The analysis of imputation techniques employed in the examined hypoglycemia prediction studies reveals that statistical imputation techniques are predominantly utilised to address missing variables. The main explanation for using them is that machine learning techniques necessitate a substantial number of data to learn patterns for predicting missing values.

In the reviewed studies, Linear Interpolation (LI) is used a maximum number of times (9) to impute missing gaps of different lengths. LI works on a simple equation in which the missing values are filled by a straight line joining the endpoints (known values). 
$$
\hat{y} = y_1 + \frac{(x_2 - x_1)(y_2 - y_1)}{x_2 - x_1}
$$
This formula calculates the estimated value ($\hat{y}$) using linear interpolation between two points $(x_1, y_1)$ and $(x_2, y_2)$, where $x$ is the input value for which the corresponding interpolated value is found.

The next commonly used interpolation technique is Spline Interpolation (SI), which has been applied four times in the studies. Instead of simultaneously fitting a single, high-degree polynomial to all the values, spline interpolation fits low-degree polynomials to small subsets. It uses the following equation \cite{b28}.

''Given a set of data points $\{(x_i, y_i)\}_{i=0}^n$, where $x_0 < x_1 < \ldots < x_n$, the spline interpolation algorithm constructs $n$ polynomials $S_i(x)$ over each subinterval $[x_i, x_{i+1}]$, such that:
\begin{itemize}
    \item $S_i(x_i) = y_i$ and $S_i(x_{i+1}) = y_{i+1}$ for $i = 0, 1, \ldots, n-1$ (interpolating conditions).
    \item $S_i'(x_{i+1}) = S_{i+1}'(x_{i+1})$ and $S_i''(x_{i+1}) = S_{i+1}''(x_{i+1})$ for $i = 0, 1, \ldots, n-2$ (continuity conditions at interior knots).
    \item $S_0''(x_0) = S_n''(x_n) = 0$ (natural spline boundary conditions)''.
\end{itemize}

K Nearest Neighbor (KNN), Mean, and Forward Fill (FF) have been used two times each. 
In the KNN method, a distance metric chooses the closest points with the missing value. Then, with those k points, the missing value is calculated by taking the mean, median, or mode \cite{KNN}. In Forward Fill, the value is filled by the closest preceding value \cite{FF}. Filling the values by mean constitutes filling the missing values by the sum divided by the number of observations of the known values of the feature \cite{FF}. 

Among the reviewed studies, Stineman Interpolation, Kalman Smoothing (KS), Polynomial Interpolation (PI), and ARMA are each used only once.

Stineman Interpolation applies cubic interpolation between two data points and preserves natural trends such as monotonicity. Kalman Smoothing is an enhancement of the Kalman filter \cite{b28} employed to estimate hidden states in a dynamic system from noisy observations, utilizing both past and future data for improved accuracy in estimations. In Hourly mean \cite{b28}, a missing value of CGM at a given timestamp is replaced by the mean of the whole CGM time series in the hour that includes the timestamp corresponding to the missing value. Polynomial Interpolation imputes the gap of missing values by a polynomial of the lowest possible degree.
The following formula gives ARMA: \cite{b27}
\[
x_t = c + \varepsilon_t + \sum_{i=1}^{p} \varphi_i x_{t-i} + \sum_{i=1}^{q} \theta_i \varepsilon_{t-i} \tag{4}
\]
Where \(x_t\) is a missing data point. \(t\) is the timestamp at which the data point is missing. \(\varphi_1, \dots, \varphi_p\), \(\theta_1, \dots, \theta_q\) are parameters, \(c\) is a constant, and the random variable \(\varepsilon_t\) is white noise.

It has been noted that specific imputation strategies are utilised for each feature during a given time interval. For small missing gaps, like approximately one to five missing values, LI and FF are implemented. SI and LI interpolate missing values of approximately an hour (12 samples in the case of glucose values). For handling larger gaps, most of the studies exclude them from the dataset, barring Md Shafiqul, Qaraqe, and Belhaouari \cite{b27}, who use previous day data for a 1-day gap and ARMA for missing values more than two days.

The potential explanations for the prevalent methodologies and categorization of imputation approaches based on temporal gaps depend upon the strength and limitation of each imputation technique. LI is used in most studies as it is the simplest way to fill in the missing values. However, it has a limitation: most physiological signals do not have a linear nature for varying time gaps. Spline interpolation uses piece-wise polynomials to impute missing values, addressing the limitations of the linear behavior inherent in linear interpolation. However, aligning with the stochastic character of physiological signals remains challenging.

In addition to the imputation strategies employed for various gaps in the reviewed studies, many analogous techniques used in the reviewed studies could be applied to specific time gaps based on similar behavior as follows:
For small gaps, KNN and hourly mean can be used. For gaps around one hour, PI, Stineman Interpolation, Makima, and Pchip Interpolation can be helpful because they use polynomial functions in various ways for imputation.  

Based on the discussion and observations, imputation techniques for a particular time gap can be concluded as follows: For small time gaps, approximately one to five missing values occurring randomly over the feature, LI, KNN, hourly mean, and FF can be used. Physiological signals exhibit linear behavior in instantaneous time, making these approaches efficient.

For longer time gaps like one hour (for example, 12 samples for glucose values), SI, PI, Stineman Interpolation, Makima, and Pchip Interpolation can be used. 
These techniques are better suited for these gaps as they use polynomials of different degrees, which are smooth and nonlinear. They can closely match the behavior of signals for a slightly more extended period.

Many imputation techniques, like several hours and days, have yet to be explored for longer time gaps. One study uses previous-day data and ARMA to fill in the missing values. One reason for this could be the lack of capabilities of statistical techniques for imputing longer time gaps. As the sensor behavior is very random over a long period, the efficiency of the statistical techniques decreases rapidly.

\begin{figure}[h!]
\centering
\begin{adjustbox}{max width=\textwidth}
\begin{tikzpicture}
\begin{axis}[
    ybar,
    symbolic x coords={
        LSTM, MSMC, CNN-LSTM, Transformer, ARIMAX, 
        XGBoost, Random Forest, QRF, DBSCAN+CNN, 
        DRNN, MLPRegressor
    },
    xtick=data,
    ylabel={Number of Uses},
    ymin=0,
    bar width=15pt,
    x tick label style={rotate=45, anchor=east},
    nodes near coords,
    enlarge x limits=0.1,
    title={Model Usage Frequency}
]
\addplot coordinates {
    (LSTM,1)
    (MSMC,1)
    (CNN-LSTM,2)
    (Transformer,2)
    (ARIMAX,1)
    (XGBoost,3)
    (Random Forest,1)
    (QRF,1)
    (DBSCAN+CNN,1)
    (DRNN,1)
    (MLPRegressor,1)
};
\end{axis}
\end{tikzpicture}
\end{adjustbox}
\caption{Prediction Models used}
\label{fig4:piechart}
\end{figure}
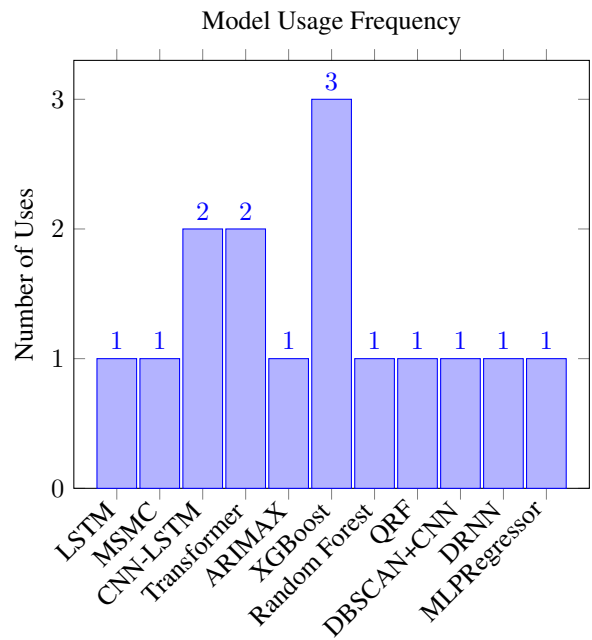

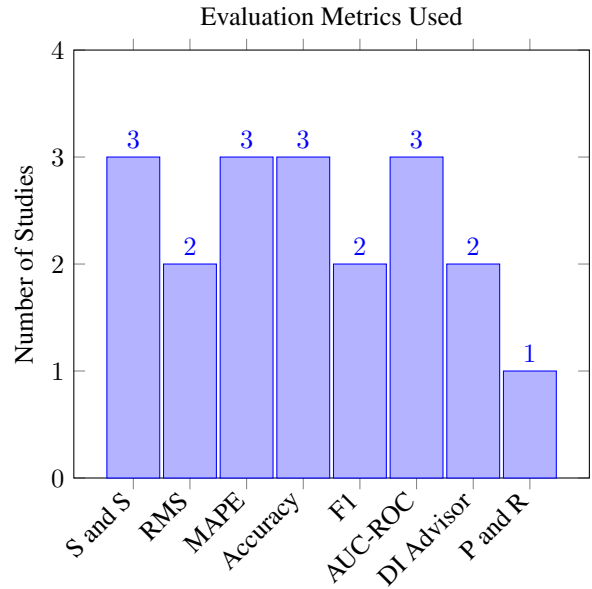
\begin{figure}[h!]
\centering
\begin{adjustbox}{max width=\textwidth}
\begin{tikzpicture}
\begin{axis}[
    ybar,
    bar width=20pt,
    ylabel={Number of Studies},
    symbolic x coords={
        S and S, 
        RMS, 
        MAPE, 
        Accuracy, 
        F1, 
        AUC-ROC, 
        DI Advisor, 
        P and R
    },
    xtick=data,
    x tick label style={rotate=45, anchor=east},
    nodes near coords,
    enlarge x limits=0.15,
    ymin=0,
    ymax=4,
    title={Evaluation Metrics Used}
]
\addplot coordinates {
    (S and S,3)
    (RMS,2)
    (MAPE,3)
    (Accuracy,3)
    (F1,2)
    (AUC-ROC,3)
    (DI Advisor,2)
    (P and R,1)
};
\end{axis}
\end{tikzpicture}
\end{adjustbox}
\caption{Evaluation Metrics Used}
\label{figure5}
\end{figure}

A comparison of the imputation techniques or a single standout technique cannot be highlighted for a specific time gap for a particular feature. This is because they use different datasets, models, and evaluation metrics.
The interpolation techniques applied in studies predicting hypoglycemia depend upon the datasets and the prediction model selected for the study. The number of studies using different machine learning prediction models and evaluation metrics are listed in figure \ref{fig4:piechart} and figure \ref{figure5}, respectively.

Based on the analysis of imputation techniques, it is concluded that multiple methods should be applied to specific gaps, and their results compared to identify the most optimal imputation approach. As seen in the previous sections, significant imputation techniques have yet to be explored to fill larger gaps. Statistical techniques may be inadequate for imputing large gaps in data due to their limited capacity to capture complex patterns. In contrast, machine learning methods offer a more effective alternative, particularly when a substantial amount of data is available. These techniques can learn underlying patterns from the existing data and apply this knowledge to more accurately impute missing values. As a result, machine learning currently represents the most promising approach among available imputation methods for handling large data gaps.
 
So, the analysis concludes that LI and SI have been used a maximum number of times for multiple features. The categorization of the imputation techniques for all the features could be made based on the review studies based on varying time gaps. Also, there is a significant research gap for imputing all the features' large gap intervals. Machine learning and deep learning techniques could be used, which will be discussed in the subsequent section.

\subsubsection {Imputation techniques used in Time Series Healthcare Datasets} \label{sec:imputation4}

Studies have been done in past years to formulate imputation techniques for time series data in the healthcare domain. The studies conducted for imputing missing data after 2018 have been included in this section. Initially, machine learning imputation techniques would be discussed followed by deep learning techniques.

\begin{enumerate}

\item \textbf{Machine Learning Imputation Techniques}

\begin{table}[h]
\caption{Machine Learning Imputation Techniques for Time series Healthcare Data}
\label{table7}
\setlength{\tabcolsep}{3pt}
\begin{tabular}{|p{95pt}|p{145pt}|}
\hline 
\textbf{Papers} & \textbf{Machine Learning Imputation Techniques Used}  \\
\hline
Luo \textit{et al.}\cite{3DMICE}  & 3D-MICE\\
\hline
Sun \textit{et al.}\cite{MICE-DA} & MICE-DA\\
\hline
Xu \textit{et al.}\cite{MD-MTS} & MD-MTS  \\
\hline
Zhang \textit{et al.} \cite{SMILES}& SMILES \\
\hline
Gao \textit{et al.} \cite{TA-DUALCV} & TA-DualCV \\
\hline 
\end{tabular}
\end{table}

Machine learning methods are used to predict or fill in missing values based on patterns learned from existing data. The process begins with training a model to understand the relationships between features that have missing values and other features in the dataset. Once the model is trained, it can predict the missing values by analysing these correlations. Table \ref{table7} shows the studies that use machine learning techniques.

One commonly used technique is Multiple Imputation by Chained Equations (MICE), which performs this process iteratively. In the first step, an initial guess is made to fill in the missing values (such as using the mean, median, or mode). Then, the model is trained to predict these missing values, and the imputed values are updated accordingly. This cycle continues until the estimates stabilize. Overall, machine learning methods utilise both cross-sectional and longitudinal dependencies to effectively impute missing values.

3D Multiple Imputations by Chained Equations (3DMICE) \cite{3DMICE,maksim} methods leverages the MICE framework for imputation. Cross-sectional imputation uses standard MICE after flattening time series data. A single-task Gaussian Process (GP) is then applied for longitudinal imputation. Both estimations are combined later using variance-informed weighted averages. 

The limitation of separate models used for longitudinal and cross-sectional variables of 3D MICE is overcome by the MICE-DA method \cite{MICE-DA},\cite{maksim}. In this, multiple imputations are proposed by chained equations with data augmentation. This method aims to augment flattened cross-sectional data with features extracted from the longitudinal data and then apply the standard MICE method. 

The Multi-Directional Multivariate Time Series (MD-MTS) \cite{MD-MTS,maksim} method for estimating missing values performs rigorous feature engineering to integrate temporal and cross-sectional features into a common imputation task. Xu \textit{et al.} \cite{MD-MTS} integrate longitudinal and cross-sectional features into a new feature set. The feature set includes the following variables: (1) variable values at the current time point, (2) chart time, (3) time stamps, (4) pre and post-values in 3-time stamps, and (5) min, max, and mean values. Later, a tree-based LightGBM regressor is trained for each variable, utilizing augmented features to impute missing values.
 
The xgbooSt MIssing vaLues In timE Series (SMILES) \cite{SMILES,maksim} method uses a similar strategy to augment longitudinal features. This methodology involves three steps. In the first step, the missing values are filled out by mean. Then, the XGboost model is trained using extracted features based on the window size to impute the missing values. Ultimately, each XGBoost model estimates values for a particular variable, similar to the MICE approach.

The core part of Time-Aware Dual-Cross-Visit (TA-DualCV) \cite{TA-DUALCV,maksim} is the dual-cross-visit imputation (DualCV), in which multivariate and temporal dependencies in cross-visit using chained equations are captured. DualCV consists of two chained equation-based modules: cross-visits feature perspective module (CFP) and cross-visits temporal perspective module (CTP). Both modules use a Gibbs sampler to impute missing values, which are combined subsequently. The time-aware augmentation mechanism then captures patient-specific correlations within each time point by applying the Gaussian Process on each patient visit. At last, the results from both components, DualCV and GP, are fused using weighted averaging.

These machine-learning imputations could fill longer gaps as they learn the pattern from the known values. 

\item \textbf{Deep Learning Imputation Techniques}

Deep learning imputation techniques refer to advanced methods used to fill in missing data in datasets using deep neural network architectures. These techniques mostly outperform traditional statistical imputation methods (like mean imputation or regression) especially when dealing with complex, nonlinear relationships in high-dimensional data. In this section, we describe the deep learning imputation techniques used in healthcare domains. 

\begin{table}
\caption{Deep Learning Imputation Techniques for Time series Healthcare Data }
\label{table8}
\setlength{\tabcolsep}{3pt}
\begin{tabular}{|p{85pt}|p{145pt}|}
\hline 
\textbf{Papers} & \textbf{ Deep Learning Imputation Techniques Used}  \\
\hline

K. Yin \textit{et al.} \cite{CATSI} & CATSI \\
\hline
Yan \textit{et al.} \cite{DETROIT} & DETROIT \\
\hline

C. Yin \textit{et al.} \cite{TAME}  &  TAME \\
\hline 

Cao \textit{et al.} \cite{BRITS} & BRITS \\
 \hline
Fortuin \textit{et al.} \cite{GP-VAE} & GP-VAE \\
 \hline
\end{tabular}
\end{table}

The deep learning techniques used for imputation are listed in Table \ref{table8}. A short description of the methods are listed below. 

Bidirectional Recurrent Imputation for Time Series (BRITS) \cite{BRITS} is a deep learning model that uses bidirectional recurrent neural networks (RNNs) to impute missing values in time series data. Unlike traditional methods that rely on assumptions about data distribution, BRITS treats missing values as learnable parameters within the RNN framework. By processing data in both forward and backward temporal directions, it captures past and future dependencies, allowing for more accurate imputation. This approach is particularly effective for datasets with complex temporal patterns and has been shown to outperform other methods in various applications, including healthcare and air quality monitoring \cite{maksim}.

The Context-Aware Time Series Imputation (CATSI) \cite{CATSI} is a context-aware mechanism for imputing missing values in clinical time series data. It comprises two main components: a context-aware recurrent imputation module and a cross-feature imputation module. The former utilizes bidirectional RNNs to model temporal dynamics, while the latter captures relationships among different features. A fusion layer combines these insights to produce the final imputed values. By incorporating a global context vector that represents the patient's overall health state, CATSI effectively handles complex missing data patterns in clinical datasets \cite{maksim}.

The deep imputer of missing temporal data (DETROIT) \cite{DETROIT} employs a fully connected neural network with eight hidden layers to impute missing values in temporal data. The model begins by initializing missing entries using methods like local mean or soft impute. It then captures temporal and cross-sectional correlations among variables to refine the imputations. DETROIT is designed to handle high-dimensional health data, making it suitable for applications where capturing intricate variable interactions over time is crucial \cite{maksim}. 

In the Gaussian Process Variational Autoencoder (GP-VAE) \cite{GP-VAE}, it combines variational autoencoders (VAEs) with Gaussian processes (GPs) to model time series data with missing values. The VAE component maps the data into a latent space, capturing underlying structures, while the GP component models temporal correlations within this space. This integration allows GP-VAE to provide smooth and coherent imputations, along with uncertainty estimates. It's particularly useful for datasets where understanding the confidence of imputations is as important as the imputations themselves \cite{maksim}.

Time-Aware Multi-modal auto-Encoder (TAME) \cite{TAME}  is designed to handle missing data in multi-modal time series, such as electronic health records that include demographics, diagnoses, medications, lab tests, and vital signs. The model integrates a time-aware attention mechanism with a bidirectional LSTM architecture to capture temporal dependencies. By incorporating multi-modal embeddings, TAME effectively learns from diverse data sources. Additionally, it employs dynamic time warping to measure patient similarity, enhancing its capability to impute missing values accurately across different modalities \cite{maksim}.
\end{enumerate}

\subsection{Discussion}
\label{sec:results}

This review has explored state-of-the-art preprocessing and imputation techniques for time series data related to hypoglycemia classification. Moving back to the research questions introduced in section \ref{sec:introduction}, it was identified that: 1) Six clinical datasets are commonly employed to predict hypoglycemia. These datasets have limitations such as insufficient data quantity, missing values of essential features at different time intervals, outliers, and disproportionate numbers of hypoglycemia cases. Furthermore, the frequency of collected values varies. The OhioT1DM dataset is the most popular multivariate dataset, and Dexcom is the most popular sensor. 
2) The most relevant features include glucose and features extracted from glucose and time, heart rate, and activity, but most studies use glucose values for hypoglycemia prediction. It is identified that each value, including BG, HR, and ACC, behaves differently over time. The change in values of different signals depends upon different parameters, which are not significantly correlated. For instance, BG levels change depending on parameters like meal intake, insulin dose, and physical exercise. After these activities, BG levels temporarily deviate but largely remain consistent throughout. Also, the heart rate readings are neither linear nor steady. The fluctuation rate of HR values is uncertain and variable. The accelerometer values magnitude lies in the range of -1 to 1; there are rapid changes in the values, but their magnitude is insignificant. Therefore, this behavior of features requires appropriate preprocessing techniques.
3) Statistical imputation techniques have been used in studies predicting hypoglycemia. It is observed that the same imputation technique is used to impute various features or time gaps. Furthermore, some studies imputed varying lengths of gaps using different imputation methods or compared different methods for the same gaps. 
The analysis of different imputation techniques concluded that linear imputation was used nine times and spline interpolation four times. K Nearest Neighbor, Mean, and Forward Fill have been used two times, and Stineman Interpolation, Kalman Smoothing, Polynomial Interpolation, and ARMA have been used single time in the studies. Only a few imputation approaches are employed for longer temporal gaps. 5) ML models from other healthcare domain like 3D MICE, MICE-DA, MD-MTS, SMILES, and TA-DualCV could be adopted for longer gaps.

Relevant challenges were identified throughout the literature review. The major constraint is the limited number, quality, and size of multivariate data collection on patients with type 1 diabetes. Consequently, more complex imputation techniques, such as deep learning methods, cannot be applied.

These constraints arises as conducting clinical trials is a difficult process and participants are required to properly wear all wearables for the study duration. In addition, technological errors and notably the human factor can lead to non-measured values, outliers or missing values.
The datasets primarily consist of three features: BG, HR, and accelerometer signals, which exhibit varying behaviors throughout time. The blood glucose values are related to these features, but then also only 5 out of 16 studies use ECG, ACC values, and other physiological signals. Reasons highlighted for not using them are subsequent missing values and insufficient data quality. 
Discussing the temporal behavior of features, BG readings exhibit minimal fluctuation over short time intervals. Therefore, the sensors are designed to record values at five-minute intervals. However, HR and ACC signals exhibit instantaneous level fluctuations for various reasons. Thus, they are recorded on a per-second or per-minute basis. As all the features have different temporal behaviors, we suggest they be imputed differently with various imputation techniques. 
Sampling all considered variables to the same frequency can lead to further data loss, overestimation, or missing values. Thus, missing values will always occur in datasets, complicating the data and necessitating extensive preprocessing and imputation methods.

The literature review lists and analyses various imputation methods used in the studies. We identify that predominantly statistical imputation techniques have imputed missing values for various lengths. They are likely used due to the smaller data quantity. Linear Imputation has been used in most studies to input length gaps for different features, followed by KNN. The possible reason for their predominant usage could be the simple architecture on which they work. We have divided imputation techniques based on their usage of length gaps, and discussion is done based on their strength and weaknesses. Different gaps are divided based on all combined features, but BG values dominate it, as many studies do not use ECG and ACC values for prediction. For ECG and ACC values, only KNN and mean values are used for imputation. One possible explanation for not using the data is that there may be more missing values and poor-quality data for these features. The classification above, derived from the evaluated research for specific missing values, comprises all integrated features; nevertheless, the primary emphasis is on the imputation of blood glucose values. As a research gap, it has been identified that more studies should supplement their data with ECG and Accelerometer values, and proper imputation techniques should be researched. Conclusively, different imputation methods per varying time gaps may increase the data quality for the same feature.

Imputation techniques for longer time intervals are insufficient as to why ML methods from other medical domains are suggested. Although the datasets based on hypoglycemia prediction are limited in data quantity, these techniques could be adopted rather than just excluding the missing values from the datasets.

From the analysis of the imputation techniques used in the reviewed studies, we propose a paradigm (see fig \ref{fig:para}) that could be followed for each feature separately. This paradigm primarily focuses on BG values because it draws inspiration from the imputation procedures employed in reviewed studies.

In this paradigm, we extract the feature that must be imputed. Before deciding on the imputation technique, a proper feature evaluation is done to study its temporal behavior. Then, the method is decided based on the data volume. Deep learning imputation techniques are suggested if the data volume is big enough. For smaller data volumes, as in the case of datasets involved in hypoglycemia prediction, a proper division of time gaps is made based on the following criteria:
\begin{itemize}
    \item Criteria 1 (C1): If the time gaps is around 1 hr (12 samples in case of Glucose values). 
    \item Criteria 2 (C2): If the missing gaps is of length upto 15 minutes.
    \item Criteria 3 (C3): If the time gaps is of length 1 day or more than 1 day.
\end{itemize}
 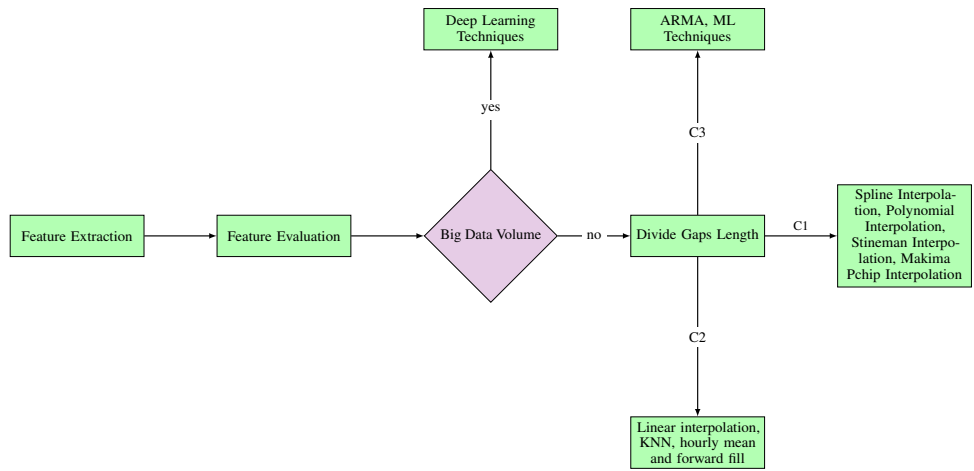
\begin{figure}
     \begin{center}
    \vspace{0.55cm}
\scalebox{0.55}{
\begin{tikzpicture}[node distance=5cm]
\node (fex) at (0,0) [select] {Feature Extraction};
\node (fe) [select, right of=fex] {Feature Evaluation};
\node (bv) [decision, right of=fe] {Big Data Volume};
\node (dlt) [select, above of=bv] {Deep Learning Techniques};
\node (dgl) [select, right of=bv] {Divide Gaps Length};
\node (tech) [select, above of=dgl] {ARMA, ML Techniques};
\node (li) [select, below of=dgl] {Linear interpolation, KNN, hourly mean and forward fill};
\node (si) [select, right of=dgl] {Spline Interpolation, Polynomial Interpolation, Stineman Interpolation, Makima Pchip Interpolation };

\draw [arrow] (fex) -- (fe) ;
\draw [arrow] (fe) -- (bv);
\draw [arrow] (bv) -- node[fill=white] {yes}(dlt);
\draw [arrow] (bv) -- node [fill=white] {no}  (dgl) ;
\draw [arrow] (dgl) -- node [fill=white] { C3}  (tech) ;
\draw [arrow] (dgl) -- node [fill=white] {C2}  (li) ;
\draw [arrow] (dgl) -- node [anchor =south] {\small C1} (si) ;
\end{tikzpicture}}
\end{center}
\caption{Paradigm of Imputation techniques}
\label{fig:para}
\end{figure}

Here, a list of imputation techniques for a specific time gap is proposed, but not a standout technique for a time gap is highlighted. The impact of preprocessing steps depends on the chosen prediction algorithm, as to why the best method cannot be highlighted in this paradigm \cite{b28}. We suggest analysing all of them for a certain time gap. Two studies reported that linear imputation is best for imputing glucose values if deep learning models are utilised as the predictor \cite{b29,b28}. 

Lastly, in most studies, the focus lies on the overall performance of the prediction model rather than on a direct comparison of the imputation methods. Acuna \textit{et al.} illustrate the actual glucose values alongside interpolated values for a specific timeframe of one subject, yet they do not report the numerical variations \cite{b28}. Therefore, it is recommended that direct comparisons of the imputation methods be conducted against both the actual and predicted samples first. Then, the performance of different imputation methods should be further compared based on their impact on the prediction model, as shown in prior studies.

Conclusively, the following research gaps are identified: 1) Multivariate data and different sensor types are not commonly applied for hypoglycemia prediction, and most studies only use glucose data. 2) Complex imputation methods based on machine learning are not applied for longer time gaps. 3) Most studies do not impute different features separately 4) Different imputation methods per feature and length of time gap are rarely investigated. 4) Imputation methods need to be thoroughly explored and compared. 

\section{Conclusion}
\label{sec:conclusion}

This review has highlighted state-of-the-art approaches for data preprocessing and imputation of glucose values and hypoglycemia features. It presents a framework of preprocessing techniques for future studies and a paradigm for employing imputation techniques in hypoglycemia prediction.

Addressing the limitations of clinical datasets poses a significant challenge in predicting hypoglycemia. Integrating data from different sensors into a unified dataset presents several issues. The limitations of clinical datasets include small sample sizes, missing values, outliers, and inconsistencies in the frequency or timestamps of certain features. These factors can lead to bias and reduce the accuracy of predictions. Missing values are a significant constraint that decreases the data quality and quantity, resulting in poorer predictive performance. To mitigate these limitations, a thorough examination of the relevant features and implementation of comprehensive preprocessing methods are required.

Analysis of features' behavior indicated that employing separate imputation techniques for different features is effective as they behave differently over time. Furthermore, applying different imputation methods for gaps of different lengths is suggested because various parameters influence the change in feature levels over time. Moreover, a lack of imputation strategies for ECG and accelerometers has been noted due to their limited use in studies forecasting hypoglycemia. It is highlighted that longer missing gaps are excluded from the datasets, and fewer imputation techniques have been explored. Therefore, we suggest using machine learning techniques from other healthcare domains to fill these larger gaps. Conclusively, a paradigm of imputation techniques for specific time gaps is proposed to be followed by future studies. The best imputation technique could not be predicted in this paradigm because accuracy depends upon the datasets and predictor models used.

Therefore, studies should thoroughly explore and compare imputation methods for separate features since the impact of these methods depends upon the chosen machine learning model, the specific features, their temporal behavior, and the dataset's characteristics. Even though missing values are addressed in many ongoing studies, the best techniques cannot be universally highlighted, as the optimal approach depends on various factors.

\end{document}